\documentclass[11pt,a4paper,english,11pt,a4paper]{article}
\usepackage[utf8x]{inputenc}
\usepackage{fancyhdr}
\usepackage{color}
\usepackage{babel}
\usepackage{booktabs}
\usepackage{amsmath}
\usepackage{graphicx}
\usepackage[authoryear]{natbib}
\usepackage[unicode=true,pdfusetitle,
 bookmarks=true,bookmarksnumbered=false,bookmarksopen=false,
 breaklinks=false,pdfborder={0 0 1},backref=false,colorlinks=true]
 {hyperref}

\makeatletter

\pdfpageheight\paperheight
\pdfpagewidth\paperwidth

\providecommand{\tabularnewline}{\\}

\@ifundefined{date}{}{\date{}}
\usepackage[T1]{fontenc}
\usepackage{babel}
\usepackage{arabtex}
\usepackage{ucs}

\usepackage[hyperref]{acl2019}
\usepackage{times}
\usepackage{latexsym}
\usepackage[super]{nth}

\usepackage{url}

\aclfinalcopy 

\title{Latent Universal Task-Specific BERT}
\author{Alon Rozental,~~ Zohar Kelrich, ~~Daniel Fleischer \\
  Amobee Inc., Tel Aviv, Israel \\ \\
   \tt alon.rozental@amobee.com \\ \tt zohar.kelrich@amobee.com\\ \tt daniel.fleischer@amobee.com }

\makeatother

\begin{document}
\maketitle 
\begin{abstract}
This paper describes a language representation model which combines
the Bidirectional Encoder Representations from Transformers (BERT)
learning mechanism described in \citet{DBLP:journals/corr/abs-1810-04805}
with a generalization of the Universal Transformer model described
in \citet{DBLP:journals/corr/abs-1807-03819}. We further improve
this model by adding a latent variable that represents the persona
and topics of interests of the writer for each training example. We
also describe a simple method to improve the usefulness of our language
representation for solving problems in a specific domain at the expense
of its ability to generalize to other fields. Finally, we release
a pre-trained language representation model for social texts that
was trained on 100 million tweets.
\end{abstract}

\section{Introduction\label{sec:Introduction}}

Several state of the art results in multiple NLP tasks were obtained
recently by relying on pre-trained language models \citep{Peters:2018,Cer:2018aa}.
Amongst the most notable of these models is BERT \citep{DBLP:journals/corr/abs-1810-04805},
which relies on the Transformer Encoder architecture \citep{DBLP:journals/corr/VaswaniSPUJGKP17}
and a unique pre-training objective known as ``masked language model''
(MLM). The objective of MLM is to predict missing vocabulary tokens
based on their context; using this objective was shown to greatly
improve the pre-training effectiveness of the model by allowing the
self-attention layers of the Transformer to attend to tokens both
before and after a missing token. An additional component of the loss
function used in BERT is the ``next sentence'' (NS) prediction loss;
the model generates a special classification vector from the text,
which supports a classifier that decides whether the text is a single
paragraph or a concatenation of two unrelated text sequences (sentences).
Another approach that yielded impressive results over the last year
focused on improving the self attention mechanism of language models
\citep{DBLP:journals/corr/abs-1807-03819,so2019evolved} and these
improvements were often shown to produce better results compared to
older Transformer based language models.

In addition to language models architectures, we explored the use
of latent variables in modeling. Latent variables have various uses
in NLP; a popular example of such use is the Latent Dirichlet Allocation
algorithm for topic modeling \citep{blei2003latent}. The latent variables
are the unobserved topics; each document is a mixture of topics and
each topic has some characteristic words. We introduce a simplified
variation of latent variables use: when modeling language, specifically
tweets, we assume each document is written by a single unknown author.
We model the author classes creating the tweets using latent variables
where each author class represents a distribution on the vocabulary
words, thus helping to predict the missing words in a tweet.

We focus on Twitter as a source for short, social, text interactions
and specifically, on predicting missing words in tweets. Regarding
the use of latent variables, we feel that our assumption of one author
per tweet is reasonably justified in this case.

Our original contribution is extending the BERT model by introducing
two new mechanisms: dynamically calculating the number of iterations
over tokens, similar to universal transformers \citep{DBLP:journals/corr/abs-1807-03819}.
Secondly, implementation of latent variables representing the different
``kind'' of authors of the tweets, increasing the accuracy of missing
word predictions, where the variables are represented in the last
layer bias terms. Finally, we present a simple technique for specializing
the training for specific tasks.

The paper is organized as following: Section \ref{sec:Latent} describes
the latent variables modeling of topics, Section \ref{sec:Universal}
describes the modifications to the BERT architecture. Section \ref{sec:Task-Specific-Preprocessing}
presents a method of specializing pre-training for specific tasks.
Sections \ref{sec:Experiment} and \ref{sec:Results} describe the
experiments and results, respectively. Finally, in Section \ref{sec:Summary}
we review and summarize.

\section{Latent Topics \label{sec:Latent}}

Knowing the topic or context of the text, as well as the interests
of the author, can be very helpful in the prediction of missing words.
For example, when reading an article about American history, the missing
word in the sentence ``He was born in \textless MISSING\textgreater ''
can be most reasonably guessed to be a city in the USA. However, when
reading an article about Polish history the previous guess would be
extremely unlikely. In language models based on the Transformer architecture
\citep{DBLP:journals/corr/VaswaniSPUJGKP17}, the context of a word
is learnt through a self-attention mechanism, where each word queries
all other words for relevant details. Such a mechanism requires a
computational complexity of $O(N^{2})$ where $N$ is the number of
words in the text; however there are relevant features that can be
extracted from the text with a few parameters and a lower complexity
with regards to the number of words in the text. Specifically, we
believe that meaningful insights about the topic of a text can be
inferred with an $O(N)$ complexity and that doing so may reduce the
inferential load from the computationally and parameter intensive
part of the model.

We suggest a mechanism to learn the topic of a text by extending the
final model's bias vector with a matrix of size $V\times L$, where
$V$ is the vocabulary size and $L$ is the latent space dimension;
in this work we take it to be $L=8$. The latent space represents
the number of possible topics, or user personas. The latent matrix
weights are learnt in the following way: we assume each example was
written under a single latent category and generate a probability
distribution over the L categories for each training example with
the probability defined as:

\begin{align}
p\left(b_{i}\right) & =0.99\cdot\underset{i\in L}{\text{softmax}}\left(\sum_{t\in T_{0}}b_{i}\left(t\right)\right)+0.01\cdot L^{-1},
\end{align}
where $p(b_{i})$ is the probability of latent category $i$ for the
sentence, $b_{i}\left(t\right)$ is the $i$th bias term for the token
$t$, $T_{0}$ is the set of unmasked tokens in the sentence and the
softmax normalization is over all $L$ latent topics. In order to
calculate the MLM loss, we construct a per-example bias vector $V_{j}$
with component $j$ representing the $j$th missing word in the example:
\begin{equation}
V_{j}=p\left(b_{i}\right)\sum_{i=1}^{L}b_{i}\left(j\right)\,.
\end{equation}
We then continue the loss calculation as described in \citet{DBLP:journals/corr/abs-1810-04805}.

In order to calculate the NS prediction loss, we first compute the
Euclidean distance and KL-divergence between the probability distributions
over latent categories of the first and second parts of the text (denoted
as sentences $A$ and $B$). Those numbers are given as additional
inputs to the classification token that was used to perform the NS
prediction. Adding the distance between the distributions improves
NS prediction in cases where it is likely that sentences $A$ and
$B$ were written by different authors.

When extracting features from text, we extract the latent bias distribution
of the text, in addition to the usual word vectors and the classification
token vector. In many cases, the different bias vectors will diverge
into human comprehensible categories. In these cases, the bias distribution
of the text can yield meaningful insights. To illustrate this, we
list in table \ref{tab:Top-tweets-table} the top tweets for each
category, i.e the tweets with the highest $p(b_{i})$ for category
$i$.
\begin{table*}[t]
\centering{}%
\begin{tabular}{cccc}
\toprule 
 & Parameters & Masked LM accuracy & Next sentence prediction accuracy\tabularnewline
\cmidrule[0.05em](r){2-2}\cmidrule[0.05em](r){3-3}\cmidrule[0.05em](r){4-4}Base
Model & 110.1M & 0.507 & 0.949\tabularnewline
Latent Model & 110.3M & \textbf{0.514} & 0.950\tabularnewline
Universal Model & 46.3M & 0.508 & \textbf{0.951}\tabularnewline
Latent Universal Model & 46.5M & 0.507 & 0.949\tabularnewline
\bottomrule
\end{tabular}\caption{\label{tab:Results}Results for all models at the end of training.
These results were obtained on a validation set of 100,000 tweets.}
\end{table*}

\section{Universal Modification\label{sec:Universal}}

In order to improve performance and reduce the number of model parameters,
we have replaced the Transformer \citep{DBLP:journals/corr/VaswaniSPUJGKP17}
model with a design, similar to the Universal Transformer \citep{DBLP:journals/corr/abs-1807-03819},
where the sequential self-attention blocks of the Transformer Encoder
are replaced with a single recurring block.

This model also incorporates an Adaptive Computation Time (ACT) mechanism,
similar to \citet{DBLP:journals/corr/Graves16}, which adjusts the
number of times the representation of each position in a sequence
is revised. Furthermore, we extended the recurrent part to have three
sequential self-attention layers, instead of one. When using this
model, an ACT ponder-time regularization loss was added to the overall
loss of the model. This loss increases linearly with the number of
revisions to each token's representation. The total number of parameters
in this model is 46.3 million, significantly less than the 110.1 million
parameters used by the original model.

\section{Task Specific Preprocessing\label{sec:Task-Specific-Preprocessing}}

In the past, there have been several successful attempts to create
``task specific'' word embeddings and classifiers \citep{tang2014learning,DBLP:journals/corr/abs-1808-08782}.
These embeddings usually outperform similar classifiers that use a
general-purpose word embedding. Recently, several state of the art
results were achieved by contextual, language model based, word embeddings
such as \citet{Cer:2018aa,Peters:2018} outperforming both general
and task-specific word embeddings.

We suggest a method of improving contextual word embeddings by adding
a class weight to each token in the vocabulary; while training a language
model, we multiply its loss by a pre-determined parameter. More concretely,
we choose a large parameter when predicting words that are known to
be relevant for the task, and a small parameter when predicting irrelevant
words. For example, emojis are important for emotion classification
and various NLP tasks so we took this parameter to be 2 for emojis.
On the other hand, we multiplied the weights of URLs—which are compressed
by Twitter in to a random sequence of characters–and Twitter \textit{mentions}—which
often look like @IlovePizza, and have very little to do with the text
of the tweet—by $0.02$.

We found that treating URLs and mentions as any other part of the
text introduces a lot of noise to our model, while trying to replace
them with unweighted special tokens such as\texttt{ \_URL\_} results
in models that only predict these frequent tokens. Overall, the weighting
step helps reduce noise and focus on the features that are important
to us.

\section{Experiment\label{sec:Experiment}}

We trained 4 model variants; we will refer to them as ``Base Model'',
''Latent Model'', ``Universal Model'' and ``Latent Universal
Model''. The Base Model was trained using the pre-training script
provided by the original BERT code. The Latent Model was augmented
with 8 latent bias layers as described in section \ref{sec:Latent}.
The Universal Model uses the model described in section \ref{sec:Universal}
and the Latent Universal Model implements both of the two improvements.

All the models were trained for 5 million batches with 20 tweets per
batch. The maximum tweet length was set to 96 tokens where the vocabulary
contained the original BERT vocabulary augmented with the 824 most
common emoticons on Twitter. In order to have the NS prediction loss,
tweets were split into 2 parts: the first sentence and the rest of
the tweet. Emoticons were considered to be a sentence splitter for
this purpose but unlike characters such as {[}. ! ?{]} the emoticon
was considered to be the first character of the new sentence instead
of the last character of the previous sentence. All models were trained
on a single Tesla V100 GPU.

\section{Results\label{sec:Results}}

After training the four aforementioned models, we compared them by
examining their accuracy on the pre-training sub-tasks. The ``Latent
Model'' achieved the highest MLM accuracy and is significantly better
than the other models (p-value \textless{} 0.001). All other tests
did not yield significantly different results in both the MLM and
NS prediction tasks. Notably, the ``Universal Model'' was able to
achieve performance equal to the base model while using less than
half of the trainable parameters. Accuracy and number of parameters
are shown in table \ref{tab:Results}.

\begin{figure*}
\begin{centering}
\includegraphics[scale=0.3]{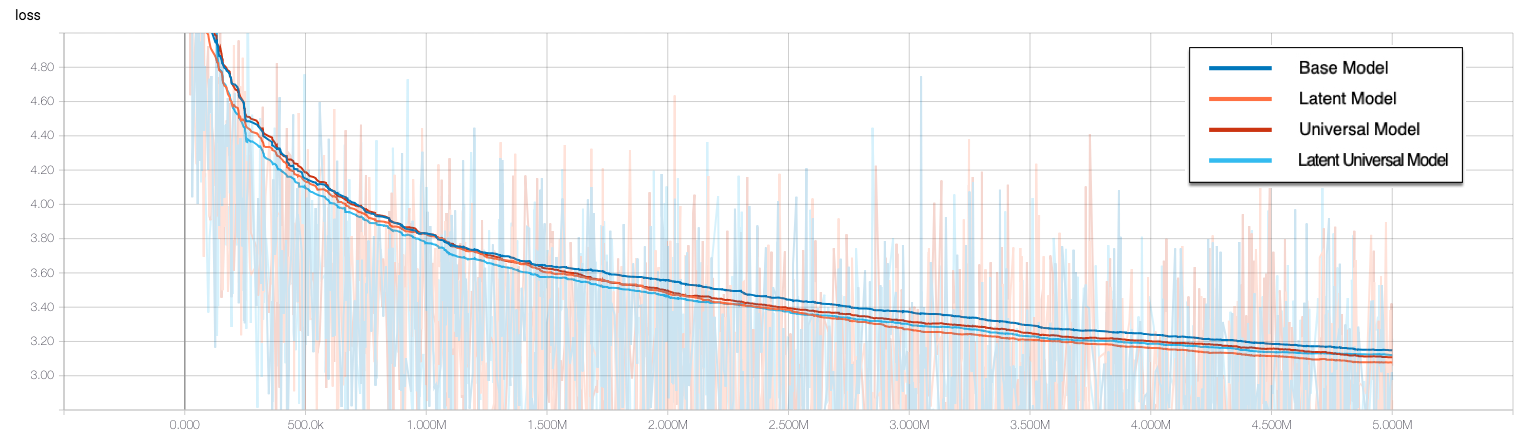}
\par\end{centering}
\caption{\label{fig:loss}This figure shows the loss over time for the different
models. Losses for the Universal Model and Latent Universal Model
also include a small ACT regularization term.}
\end{figure*}

\begin{figure*}
\begin{centering}
\includegraphics[scale=0.3]{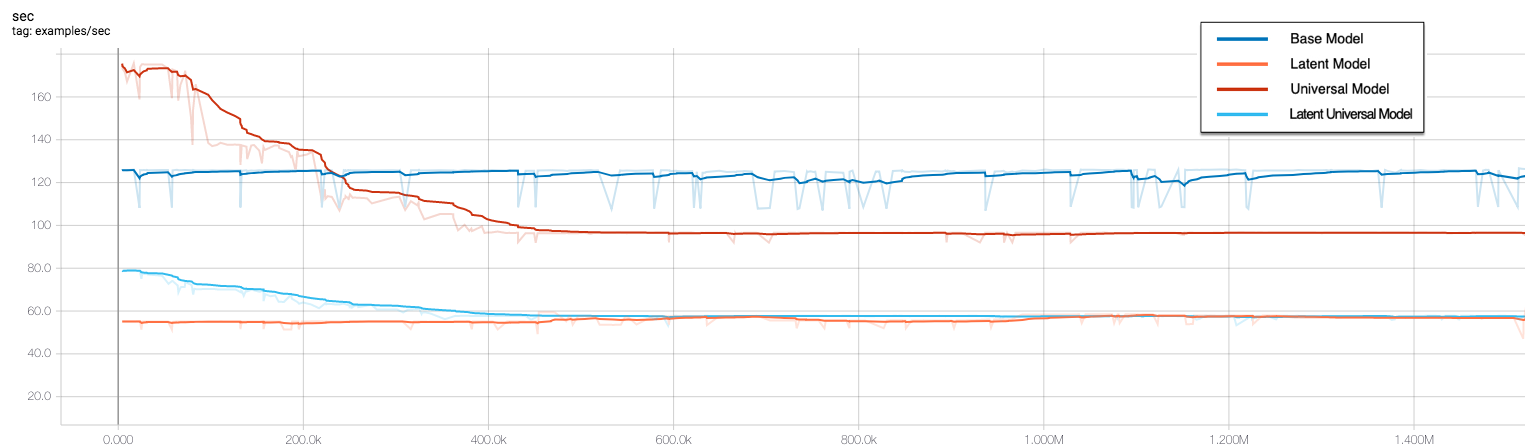}
\par\end{centering}
\caption{\label{fig:speed}The figure shows the running speed of the different
models. The recurrent models become slower over time as they learn
to repeat the self attention step of the model more times, though
this tendency is weaker when having latent bias variables.}
\end{figure*}

Another noticeable finding is that the recurrent models (Universal
and Latent Universal) become slower over time as they learn to perform
more repetitions of the recurrent part of the model. This is a result
of having a regularization factor in these models for the number of
recurrent repetitions; as the recurrent part gets better, it becomes
more worthwhile to ``pay'' the regularization cost. Interestingly,
adding the aforementioned latent bias variables weakens this tendency.
While the Universal model ends up \textasciitilde 25\% slower than
the corresponding Base model, the Latent Universal model is just as
fast as the corresponding Latent model, showcasing that adding the
latent bias variables alleviates some of the inferential load from
the self-attention part of the model. For loss over time see figure
\ref{fig:loss}; for examples rate over time see figure \ref{fig:speed}.

\section{Summary and Conclusions \label{sec:Summary}}

In this paper we described a system that combines the loss function
derived from \citet{DBLP:journals/corr/abs-1810-04805} with a recurrent
variant of the Transformer architecture, called Universal Transformer.
In addition, we modeled independent authors as latent variables by
expanding the bias term and modifying the loss. We have shown that
making the described changes can improve both the words prediction
accuracy and reduce the complexity of the model when preforming the
pre-training phase on social textual data.

The code used to produce the above results and the trained Latent
model can be found in\footnote{\href{https://s3.amazonaws.com/amobee-research-public/language-model/latent_5M_bert.zip}{https://s3.amazonaws.com/amobee-research-public/language-model/latent\_5M\_bert.zip}}.

\bibliographystyle{acl_natbib}
\bibliography{bert}

\begin{table*}
\begin{centering}
\includegraphics[viewport=20bp 170bp 850bp 800bp,scale=0.85]{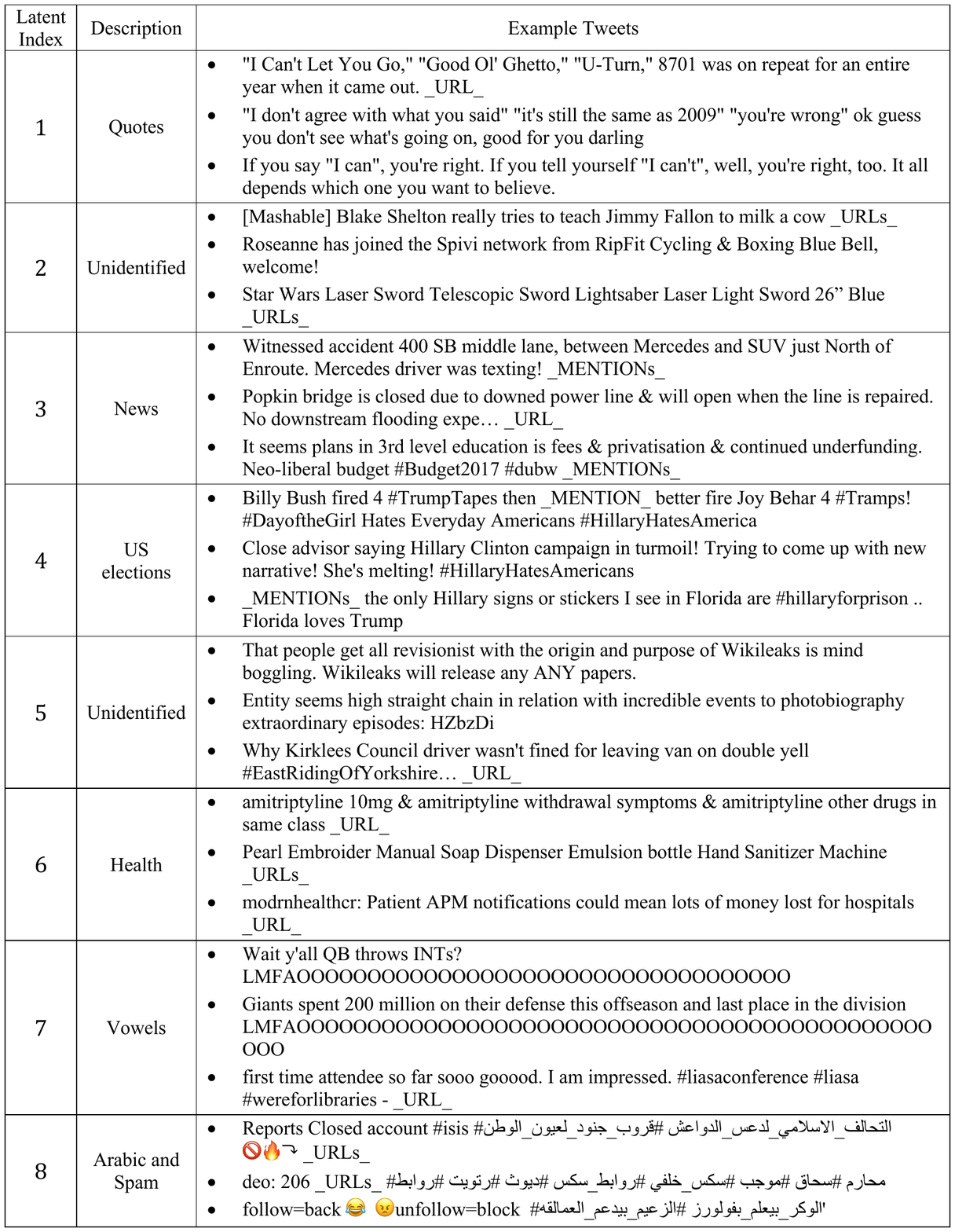}
\par\end{centering}
\caption{\label{tab:Top-tweets-table}The table contains the top tweets for
each latent category and a description of the category.  }
\end{table*}

\end{document}